\documentclass{article}
\usepackage{ijcai20}

\usepackage{times}
\usepackage{soul}
\usepackage{url}
\usepackage[hidelinks]{hyperref}
\usepackage[utf8]{inputenc}
\usepackage[small]{caption}
\usepackage{graphicx}
\usepackage{amsmath}
\usepackage{amsthm}
\usepackage{booktabs}
\usepackage{algorithm}
\usepackage{algorithmic}

\usepackage{hyperref}       
\usepackage{url}            
\usepackage{booktabs}       
\usepackage{amsfonts}       
\usepackage{nicefrac}       
\usepackage{microtype}      
\usepackage{amsmath}
\usepackage{bbm}
\usepackage{graphicx}
\usepackage{subcaption}
\usepackage{algorithmic,algorithm}
\usepackage{enumitem}
\usepackage{wrapfig}

\newcommand{\paratitle}[1]{\vspace{1.5ex}\noindent \textbf{#1}}
\newtheorem{prop}{Proposition}

\author{
Kwei-Herng Lai\footnote{These two authors contributed equally in this work}\and
Daochen Zha$^*$\and
Yuening Li\and
Xia Hu
\affiliations
Department of Computer Science and Engineering, Texas A\&M University\\
\emails
\{khlai037, daochen.zha, yueningl, xiahu\}@tamu.edu
}
\title{Dual Policy Distillation}

\begin{document}
\maketitle
\begin{abstract}
  Policy distillation, which transfers a teacher policy to a student policy has achieved great success in challenging tasks of deep reinforcement learning. 
  This teacher-student framework requires a well-trained teacher model which is computationally expensive. Moreover, the performance of the student model could be limited by the teacher model if the teacher model is not optimal. In the light of collaborative learning, we study the feasibility of involving joint intellectual efforts from diverse perspectives of student models. In this work, we introduce \emph{dual policy distillation}~(DPD), a student-student framework in which two learners operate on the same environment to explore different perspectives of the environment and extract knowledge from each other to enhance their learning. The key challenge in developing this dual learning framework is to identify the beneficial knowledge from the peer learner for contemporary learning-based reinforcement learning algorithms, since it is unclear whether the knowledge distilled from an imperfect and noisy peer learner would be helpful. To address the challenge, we theoretically justify that distilling knowledge from a peer learner will lead to policy improvement and propose a disadvantageous distillation strategy based on the theoretical results. The conducted experiments on several continuous control tasks show that the proposed framework achieves superior performance with a learning-based agent and function approximation without the use of expensive teacher models.
  
\end{abstract}

\section{Introduction}
\vspace{-0.05cm}
Reinforcement learning~(RL), especially deep reinforcement learning has achieved great success in various domains~\cite{sutton2018reinforcement}, ranging from robotic control~\cite{levine2016end}, perfect information games~\cite{silver2017mastering} to imperfect information games~\cite{zha2019rlcard}. However, it usually requires a large number of interactions with the environment to obtain high-level performance~\cite{salimans2017evolution}. Recently, works have been proposed to study how we can transfer knowledge from one or more teacher models to a student model so that we can train an agent based on a pre-trained expert model~\cite{rusu2015policy,czarnecki2019distilling}. One of the simple yet effective techniques is called policy distillation~\cite{rusu2015policy}, which uses supervised regression to train a student model to produce the same output distribution as the teacher model. Policy distillation has achieved great success and led to stronger performance in challenging domains~\cite{teh2017distral,yin2017knowledge}. Unfortunately, it is computationally expensive to obtain a teacher policy since a well-performed pre-trained model is often not available. In addition, the performance of the student model could be restrained by the teacher model if the teacher model is sub-optimal.

In cognitive psychology, collaborative learning illustrates a situation in which a group of students works together to search for solutions~\cite{dillenbourg1999collaborative}. Different from the traditional teacher-student relationship where students non-interactively receive information from teachers, collaborative learning involves joint intellectual efforts from diverse perspectives of students~\cite{smith1992collaborative}. Motivated by this, we study the feasibility of collaborative learning on student policies without the use of pre-trained teacher models, and the methodology of extracting beneficial knowledge from a peer policy to accelerate learning, analogous to the human ability to learn from others.

In this paper, we introduce \emph{dual policy distillation}~(DPD), a student-student framework in which two policies operate on the same environment and extract knowledge from each other to benefit their learning. There are mainly two challenges in developing such dual learning framework upon contemporary learning-based RL agents. First, different from conventional policy distillation, which uses an expert policy, both policies in DPD are imperfect and may generate noisy outputs in the training process. It is unclear whether the regression to these noisy data is helpful. Second, contemporary RL algorithms usually use function approximators to approximate the policy function and the value function. The inaccurate estimations of the functions make it challenging to design practical algorithms that can be combined with learning-based agents.

To address the challenges above, apart from the original policy, we introduce another policy which is simultaneously trained in the same environment with different initialization. Each of the two policies iteratively optimizes its own RL objective and updates a distillation objective which extracts knowledge from the other peer policy. One may find it surprising that the proposed framework tries to simultaneously encourage the uniqueness of the policy, and keep the two polices close with distillation objective. In the following sections, we demonstrate that this method is able to balance exploration and exploitation through parallelization and distillation, respectively, with two nice properties: (1) it does not require an expert policy as teacher signals in the sense that the two student policies explore different aspects of the environment and share knowledge with each other; (2) distilling knowledge from a peer policy has theoretical policy improvement and it can achieve satisfactory performance with a learning-based agent and function approximation in our empirical results.

Through addressing the challenges, in this paper, we make the following contributions:

\begin{itemize}[wide=0pt, leftmargin=\dimexpr\labelwidth + 2\labelsep\relax]
    \item We introduce dual policy distillation~(DPD), a student-student framework in which two polices extract beneficial knowledge from each other to help their learning.
    \item We provide a theoretical justification of the policy improvement of DPD. We show that in the ideal case, by distilling a hypothetical hybrid policy, each of the policies has guaranteed policy improvement.
    \item We propose a practical algorithm\footnote{\url{https://github.com/datamllab/dual-policy-distillation}} based on our theoretical results. The algorithm uses a disadvantageous policy distillation strategy which prioritizes the distillation at disadvantage states and pushes each of the two policies towards the optimal policy. Experiments on several continuous control tasks demonstrate that the proposed DPD significantly enhances each of the two policies. 
\end{itemize}

\section{Preliminaries}
\vspace{-0.05cm}
In this section, we introduce reinforcement learning~(RL), the background of policy distillation, and the notations used in this paper.

In the following, we consider standard reinforcement learning which is denoted by a sextuple $(\mathcal{S}, \mathcal{A}, \mathcal{P}_T, \mathcal{R}, \gamma, p_0)$, where $\mathcal{S}$ is the set of states, $\mathcal{A}$ is the set of actions, $\mathcal{P}_T: \mathcal{S} \times \mathcal{A} \to \mathcal{S}$ is the state transition function, $\mathcal{R}: \mathcal{S} \times \mathcal{A} \times \mathcal{S} \to \mathbb{R}$ is the reward function, $\gamma \in (0, 1)$ is the discount factor, and $p_0$ is the distribution of the initial state. The interactions with the environment can be formalized as a Markov decision process: at each timestep $t$, an agent takes an action $a_t \in \mathcal{A}$ at state $s_t \in \mathcal{S}$ and observes the next state $s_{t+1}$ with a reward signal $r_t$. This results in a trajectory $\tau$ which consists of a sequence of triplets of states, actions and rewards, i.e., $\tau = \{(s_t, a_t, r_t)\}_{t=1,...,T}$, where $T$ is the terminal timestep. The objective of RL algorithm is to learn a policy $\pi: \mathcal{S} \to \mathcal{A}$ that maximizes the cumulative reward $R = \mathbb{E}[\sum_{t=1}^{\infty}\gamma^t r_t]$.

We use standard definitions of value function, state-action value function, and advantage function of a policy $\pi$, i.e., $V^\pi(s_t) = \mathbb{E}_{a_t,s_{t+1},...|\pi}[\sum_0^{\infty}\gamma^l r_{t+1}]$, $Q^\pi(s_t, a_t) = \mathbb{E}_{s_{t+1},a_{t+1},...|\pi}[\sum_{t=0}^{\infty}\gamma^l r_{t+1}]$, and $A^\pi(s_t, a_t) = Q^\pi(s_t, a_t) - V^\pi(s_t)$, where $a_t,s_{t+1},...|\pi$ and $s_{t+1},a_{t+1},...|\pi$ denote the resulting trajectories from the environment if we follow policy $\pi$. We use $\rho_\pi(s)$ to denote discounted visitation frequencies of state $s$, that is , $\rho_\pi(s) = \sum_{t=0}^{\infty}\gamma^t p(s_t=s)$, where $p(s_t=s)$ is the probability of $s$ being visited at timestep $t$.

Policy distillation is a simple yet effective method of transferring knowledge from one or more action policies to an untrained network. We denote $\widetilde{\pi}$ as a teacher policy, i.e., a trained model that can generate expert data, and $\pi_\theta$ as an untrained parametric student policy. Policy distillation trains the student policy by conducting regression to the teacher policy, i.e., minimizing the following objective:
\begin{equation}
\label{label:eq1}
    \mathcal{J} = \mathbb{E}_{s \sim \widetilde{\pi}} [D(\pi_\theta(\cdot|s), \widetilde{\pi}(\cdot|s))]
\end{equation}
where $s \sim \widetilde{\pi}$ means $s$ follows the distribution of $\rho_\pi(\cdot)$, and $D(\cdot, \cdot)$ is a kind of distance metric. The above description is a general form of policy distillation. There are multiple choices for the distance metrics, such as mean square error, KL divergence or log-likelihood loss~\cite{rusu2015policy}. In this paper, we use mean square error for its simplicity.

Note that, in our presented algorithm, we use $\widetilde{\pi}_\phi$ to denote a trainable peer policy with parameters $\phi$. $\pi_\theta$ and $\widetilde{\pi}_\phi$ are both student models and are trained interactively.

\section{Dual Policy Distillation}
\begin{figure}[]
    \centering
    \includegraphics[width=6.5cm]{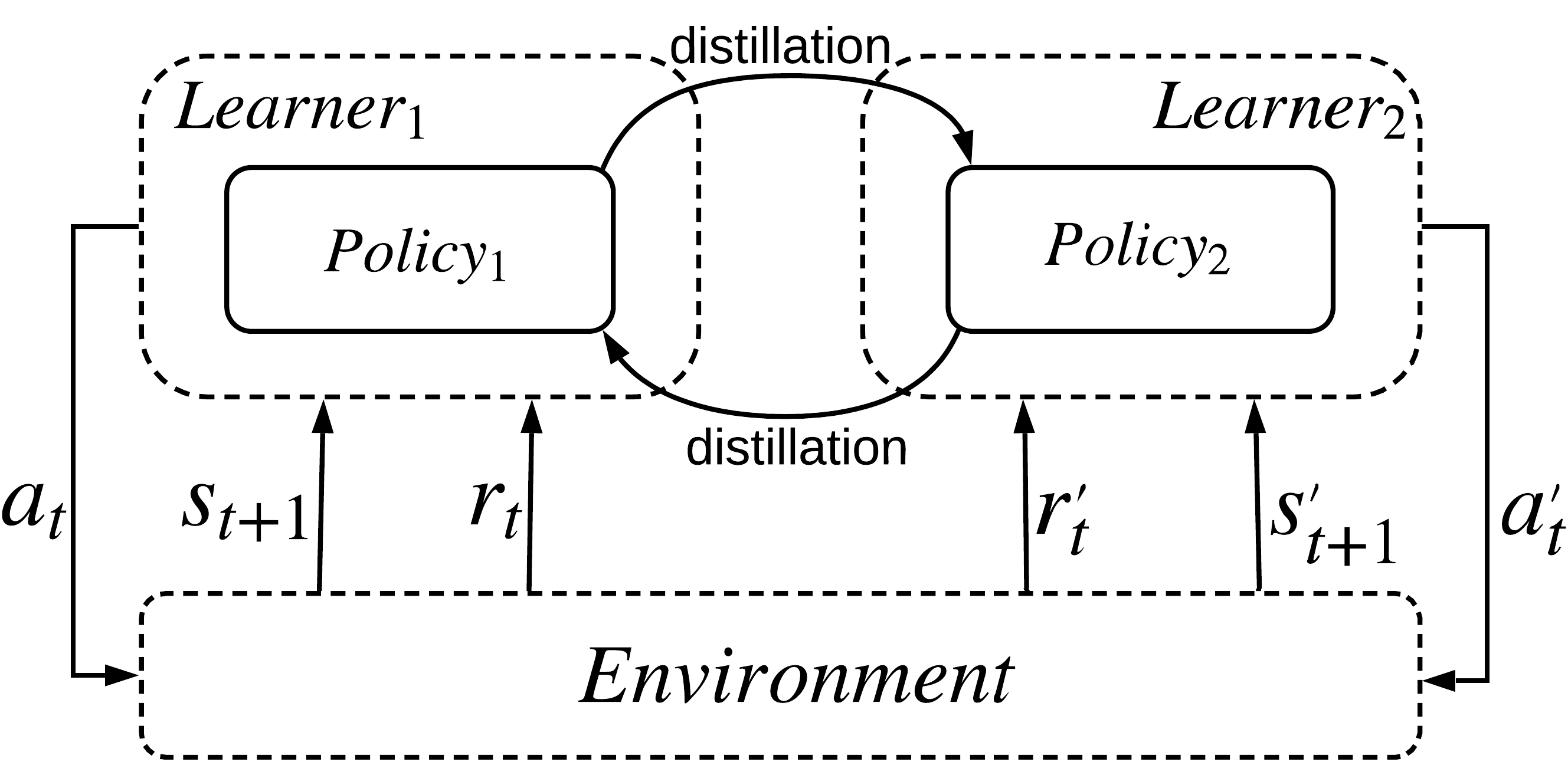}\
    \caption{An overview of dual policy distillation~(DPD) framework. The two learners interact with the same environment with different initialization. In each iteration, each of the two learners updates two objectives: the RL objective which optimizes the cumulative reward on the environment, and the distillation objective which conducts regression to its peer policy.}
    \label{fig:overview}
\vspace{-3pt}
\end{figure}

\vspace{-0.05cm}
In this section, we present dual policy distillation~(DPD), a framework that enables knowledge transfer between two student policies operating on the same environment. We consider two policies denoted as $\pi$ and $\widetilde{\pi}$ respectively. We first theoretically justify that extracting knowledge from a peer policy will lead to policy improvement through a view of hypothetical hybrid policy. Then based on our theoretical results, we present a disadvantageous policy distillation objective that can be combined with learning-based RL algorithms. Figure~\ref{fig:overview} shows an overview of the proposed framework.

\vspace{-3pt}
\subsection{A View of Hypothetical Hybrid Policy}
\vspace{-0.05cm}
We first justify that $\pi$ and $\widetilde{\pi}$ are theoretically complementary, and thus transferring knowledge between two policies will lead to policy improvement for both $\pi$ and $\widetilde{\pi}$. Consider a hypothetical hybrid policy:
\begin{equation}
\label{label:eq2}
\pi^{hypo}(\cdot|s)=\left\{
\begin{array}{rcl}
\widetilde{\pi}(\cdot|s)       &      & \xi^{\widetilde{\pi}}(s) > 0\\
\pi(\cdot|s)     &      & otherwise\\
\end{array} \right.
\end{equation}
where $\xi^{\widetilde{\pi}}(s) = V^{\widetilde{\pi}}(s) - V^{\pi}(s)$ represents the advantage of $\widetilde{\pi}$ over $\pi$ at state $s$. That is, the hypothetical policy $\pi^{hypo}$ selectively follows one of $\pi$ and $\widetilde{\pi}$ depending on which policy has larger expected discounted reward at the state. In Proposition~\ref{label:prop1}, we give that in ideal case the hypothetical hybrid policy is an improved policy compared to $\pi$ or $\widetilde{\pi}$.
\begin{prop}
\label{label:prop1}
For the hypothetical hybrid policy $\pi^{hypo}$ defined in Eq.~\ref{label:eq2}, we have $\forall s \in \mathcal{S}$, $V^{\pi^{hypo}}(s) \ge V^{\pi}(s)$ and $V^{\pi^{hypo}}(s) \ge V^{\widetilde{\pi}}(s)$.
\end{prop}
\textbf{Proof:} Considering a state $s \in \mathcal{S}$, we prove that $V^{\pi^{hypo}}(s) \ge V^{\pi}(s)$. Define advantage policy at $s$:
\begin{equation}
\pi^{adv}_s=\left\{
\begin{array}{rcl}
\widetilde{\pi}       &      & \xi^{\widetilde{\pi}}(s) > 0\\
\pi     &      & otherwise.\\
\end{array} \right.
\end{equation}
That is, $\pi^{adv}_s$ is one of $\pi$ and $\widetilde{\pi}$ such that it has higher value at state $s$. Note that whether $\pi^{adv}_s(\cdot|s')$ is $\pi$ or $\widetilde{\pi}$ depends on $s$ instead of $s'$, where $s'$ is a different state. It is straightforward to have
\begin{equation}
    V^{\pi^{adv}_{s}} (s) \ge V^\pi (s),
\end{equation}
\vspace{-5pt}
\begin{equation}
    V^{\pi^{adv}_{s}} (s) \ge V^{\pi^{adv}_{s'}} (s),
\end{equation}
where $s'$ is another state and $s \neq s'$.

\vspace{-1.5pt}

We now consider an arbitrary state $s_i \in S$ and denote $s_{i+1}$ as the next state. Define
\begin{equation}
V_n(s_i)=\left\{
\begin{array}{lcl}
\mathbb{E}_{(s_{i+1}, r_i) \sim \pi^{hypo}(s_i)} [r_i + \gamma V_{n-1} (s_{i+1})] & n \ge 1\\
V^{\pi^{adv}_{s_i}}(s_i)     &  n = 0,\\
\end{array} \right.
\end{equation}
where $n$ is a positive integer. That is, the value in state $s_i$ if we follow $\pi^{hypo}$ for the first $n$ steps and follow $\pi^{adv}_{s_{i+n}}$ afterwards.

\vspace{-1.5pt}

When $n=0$ we have $V_{0} (s_{i}) = V^{\pi^{adv}_{s_i}}(s_i)$. We first prove $\forall s_i \in S$, $V_1(s_i) \ge V_0(s_i)$.
\begin{equation}
\begin{split}
V_{1}(s_i) & = \mathbb{E}_{(s_{i+1}, r_i) \sim \pi^{hypo}(s_i)} [r_i + \gamma V_{0} (s_{i+1})]\\
& = \sum_{s_{i+1}, r_i} p^{\pi^{hypo}}(s_{i+1}, r_i | s_i)[r_i + \gamma V_{0} (s_{i+1})]\\
& = \sum_{s_{i+1}, r_i} p^{\pi^{hypo}}(s_{i+1}, r_i | s_i)[r_i + \gamma V^{\pi^{adv}_{s_{i+1}}}(s_{i+1})]\\
& \ge \sum_{s_{i+1}, r_i} p^{\pi^{hypo}}(s_{i+1}, r_i | s_i)[r_i + \gamma V^{\pi^{adv}_{s_i}}(s_{i+1})]\\
& = \sum_{s_{i+1}, r_i} p^{\pi^{adv}_{s_i}}(s_{i+1}, r_i | s_i)[r_i + \gamma V^{\pi^{adv}_{s_i}}(s_{i+1})]\\
& = V^{\pi^{adv}_{s_i}}(s_i) \\
& = V_0(s_i).
\end{split}
\end{equation}

Based on the equations above, when $k \ge 1$, given that $\forall s_i \in S$, $V_k(s_i) \ge V_{k-1}(s_i)$, we have
\begin{equation}
\begin{split}
V_{k+1}(s_i) & = \mathbb{E}_{(s_{i+1}, r_i) \sim \pi^{hypo}(s_i)} [r_i + \gamma V_{k} (s_{i+1})]\\
& = \sum_{s_{i+1}, r_i} p^{\pi^{hypo}}(s_{i+1}, r_i | s_i)[r_i + \gamma V_{k} (s_{i+1})]\\
& \ge \sum_{s_{i+1}, r_i} p^{\pi^{hypo}}(s_{i+1}, r_i | s_i)[r_i + \gamma V_{k-1} (s_{i+1})]\\
&= \mathbb{E}_{(s_{i+1}, r_i) \sim \pi^{hypo}(s_i)} [r_i + \gamma V_{k-1} (s_{i+1})]\\
& = V_k(s_i).
\end{split}
\vspace{-50pt}
\end{equation}
By induction, we can conclude that $\forall n \ge 0$, $\forall s \in \mathcal{S}$, $V_n(s) \ge V^{\pi^{adv}_{s}}(s) \ge V^{\pi}(s)$. Thus, $\forall s \in \mathcal{S}$, $V^{\pi^{hypo}}(s) \ge V^{\pi}(s)$. Similarly, we have $\forall s \in \mathcal{S}$, $V^{\pi^{hypo}}(s) \ge V^{\widetilde{\pi}}(s)$. The above proof is also applicable in continuous space if we replace sum operation with integration.

Directly following the well-known policy improvement theorem~\cite{sutton2018reinforcement}, Proposition~\ref{label:prop1} suggests that, the hypothetical hybrid policy $\pi^{hypo}$ defined in Eq.~\ref{label:eq2} is at least as good as $\pi$ and $\widetilde{\pi}$. If $V^{\pi}(s) > V^{\widetilde{\pi}}(s)$ at some states and $V^{\pi}(s') < V^{\widetilde{\pi}}(s')$ at some other states, $\pi^{hypo}$ will be strictly better than $\pi$ and $\widetilde{\pi}$. Our empirical observation also supports this intuition~(see Figure~\ref{fig:analysis}). Thus, it will lead to theoretical policy improvement if we let $\pi$ and $\widetilde{\pi}$ conduct regression to the hypothetical hybrid policy $\pi^{hypo}$. 

\subsection{Disadvantageous Policy Distillation}
\vspace{-0.05cm}
Based on the above theoretical results, we introduce a practical dual distillation strategy which can be combined with learning-based RL agents. 

For a practical algorithm, rather than building the hypothetical hybrid policy $\pi^{hypo}$ at every step, we prefer to use an objective to train the policy. In Proposition~\ref{label:prop2}, we give that under mild conditions, the distillation of $\pi^{hypo}$ is equivalent to minimizing a simple objective.  
\begin{prop}
\label{label:prop2}
The distillation to $\pi$ from the hypothetical hybrid policy $\pi^{hypo}$ defined in Eq.~\ref{label:eq2} is equivalent to minimizing the following objective:
\begin{equation}
\label{label:eq3}
    \mathcal{J} = \mathbb{E}_{s \sim \widetilde{\pi}} [D(\pi(\cdot|s), \widetilde{\pi}(\cdot|s)) \mathbbm{1}(\xi^{\widetilde{\pi}}(s) > 0)],
\end{equation}
where $\mathbbm{1} (\cdot)$ is the indicator function and $D(\cdot, \cdot)$ is denoted as the distance metric.
\end{prop}
\textbf{Proof:} Since the two policies $\pi$ and $\widetilde{\pi}$ have similar state visiting frequency, the distillation to $\pi^{hypo}$ defined in Eq.~\ref{label:eq1} can be rewritten as follow:
\begin{equation}
\begin{split}
& \mathbb{E}_{s \sim \pi^{hypo}} [D(\pi(\cdot|s), \pi^{hypo}(\cdot|s))] \\
 = & \sum_{s \sim \rho_{\pi^{hypo}}(\cdot)} \!\!\!\!\!\!\!\!  D(\pi(\cdot|s), \pi^{hypo}(\cdot|s))\\ = & \sum_{s \sim \rho_{\pi^{hypo}}(\cdot);\xi^{\widetilde{\pi}}(s) > 0} \!\!\!\!\!\!\!\!\!\!\!\!\!\!\!\!\! D(\pi(\cdot|s), \widetilde{\pi}(\cdot|s)) +\!\!\!\!\!\!\!\!\!\!\!\!\!\! \sum_{s \sim \rho_{\pi^{hypo}}(\cdot); \xi^{\widetilde{\pi}}(s) \le 0} \!\!\!\!\!\!\!\!\!\!\!\!\!\!\!\!\! D(\pi(\cdot|s), \pi(\cdot|s))\\
 =& \sum_{s \sim \rho_{\widetilde{\pi}}(\cdot);\xi^{\widetilde{\pi}}(s) > 0} \!\!\!\!\!\!\!\!\!\!\!\! D(\pi(\cdot|s), \widetilde{\pi}(\cdot|s)) + \!\!\!\!\!\!\!\! \sum_{s \sim \rho_{\widetilde{\pi}}(\cdot);\xi^{\widetilde{\pi}}(s) \le 0} \!\!\!\!\!\!\!\!\!\!\!\! D(\pi(\cdot|s), \pi(\cdot|s))\\
 = & \mathbb{E}_{s \sim \widetilde{\pi}} [D(\pi(\cdot|s), \widetilde{\pi}(\cdot|s)) \mathbbm{1}(\xi^{\widetilde{\pi}}(s) > 0)]\\
\end{split}
\end{equation}

The difference between $\rho_{\pi}(\cdot)$ and $\rho_{\widetilde{\pi}}(\cdot)$ can be ignored in our dual learning setting, since the dual distillation will push the two policies to perform similar actions and hence will result in similar state visiting frequencies. Our empirical observations also support this assumption in that both the learning curves and the outputs of the policy network are very similar during the learning process~(see Figure~\ref{fig:analysis}). The result of Eq.~\ref{label:eq3} suggests a simple intuition: the states at which the peer policy is advantageous will be more helpful in distillation; otherwise, maintaining the current policy at the state would be a better choice. We call this strategy \emph{disadvantageous policy distillation} because it prioritizes the distillation of the states at which the current policy is more disadvantageous than the peer policy.

\begin{algorithm}[t]
\caption{DPD: dual policy distillation}
\label{label:alg1}
\begin{algorithmic}[1]
\STATE \textbf{Input:} policy $\pi_\theta$, peer policy $\widetilde{\pi}_\phi$, and maximum iteration number $M$
\FOR{iteration = $1, M$}
    \STATE Execute $\pi_\theta$ to generate trajectories and save them into buffer $B_\pi$
    \STATE Update $\pi_\theta$ based on its RL objective
    \STATE Update $\pi_\theta$ based on Eq.~\ref{label:eq4}
    \STATE Execute $\widetilde{\pi}_\phi$ to generate trajectories and save them into buffer $B_{\widetilde{\pi}}$
    \STATE Update $\widetilde{\pi}_\phi$ based on its RL objective
    \STATE Update $\widetilde{\pi}_\phi$ based on Eq.~\ref{label:eq5}
\ENDFOR
\end{algorithmic}
\end{algorithm}
For now, we ignore the estimation error for the value functions which are usually approximated by deep neural networks. However, in the approximate setting, we have to consider the inaccurate estimations of value functions, which makes it difficult to optimize Eq.~\ref{label:eq3}. In our preliminary experiments, we observed that directly update the policies based on Eq.~\ref{label:eq3} will misclassify many states and lead to sub-optimal performances. Therefore, we propose to soften Eq.~\ref{label:eq3} and introduce weighted objectives $\mathcal{J}^w_{\pi_\theta}(\theta)$ and $\mathcal{J}^w_{\widetilde{\pi}_\phi}(\phi)$ to update parametric policies $\pi_\theta$ and $\widetilde{\pi}_\phi$:
\begin{equation}
\label{label:eq4}
    \mathcal{J}^w_{\pi_\theta}(\theta) = \mathbb{E}_{s \sim \widetilde{\pi}_\phi} [D(\pi_\theta(\cdot|s), \widetilde{\pi}_\phi(\cdot|s)) \exp (\alpha \xi^{\widetilde{\pi}_\phi}(s))],
\end{equation}
\vspace{-10pt}
\begin{equation}
\label{label:eq5}
    \mathcal{J}^w_{\widetilde{\pi}_\phi}(\phi) = \mathbb{E}_{s \sim \pi_\theta} [D(\widetilde{\pi}_\phi(\cdot|s), \pi_\theta(\cdot|s)) \exp (\alpha \xi^{\pi_\theta}(s))],
\end{equation}
where $\exp(\alpha \xi^{\widetilde{\pi}_\phi}(s))$ and $\exp(\alpha \xi^{\pi_\theta}(s))$ are confidence scores, and $\alpha$ controls the confidence level which should be chosen depending on how accurate the value function estimation is.

We now describe how we combine the objectives in Eq.~\ref{label:eq4}~and~\ref{label:eq5} with a learning-based agent. Given two parametric policies $\pi_\theta$ and $\widetilde{\pi}_\phi$ operating on the same environment, we use an alternating update framework as follows. In the first step, policy $\pi_\theta$ is updated based on its own RL objective. In the second step, $\pi_\theta$ is updated based on the distillation objective. Specifically, we sample a mini-batch of transitions from the buffer of $\widetilde{\pi}_\phi$ and compute the advantage $\xi^{\widetilde{\pi}_\phi}(s)$ for each sampled state and update the policy based on the distillation objective. We then do the same updates to its peer policy $\widetilde{\pi}_\phi$. As a result, each policy learns to optimize its RL objective and simultaneously extracts useful knowledge from its peer policy to enhance itself. The algorithm is summarized in  Algorithm~\ref{label:alg1}.

\subsection{Connection to Value Iteration}
\vspace{-0.05cm}
In this subsection, we justify that the proposed distillation objective will lead to similar effects as by classical value iteration~\cite{sutton2018reinforcement}. Let $\pi^*$ be an optimal policy and $V^{\pi^*}(s)$ be the optimal values, i.e., the expected discounted future rewards if we start from $s$ and follow the optimal policy. Value iteration method iteratively updates the state values as follow:
\begin{equation}
\label{label:eq6}
    V_{i+1}(s) \leftarrow \max_{a} \sum_{s'} \mathcal{P}_T (s' | s, a) [\mathcal{R}(s, a, s') + V_i(s')],
\end{equation}
where $V_{i}(\cdot)$ and $V_{i+1}(\cdot)$ are the values of the current step and the next step respectively. The intuition of this update rule is to compute the maximum value by choosing the most valuable action in each iteration to push the values towards and finally converge to the optimal values. Define $\mathcal{S}_{\widetilde{\pi}}$ as the set of disadvantage states, i.e, $S_{\widetilde{\pi}} = \{s| V^{\widetilde{\pi}}(s) > V^{\pi}(s)\}$. Then it is straightforward to have
\begin{equation}
\label{label:eq7}
    V^{\pi^*} (s) \ge V^{\widetilde{\pi}}(s) > V^{\pi}(s), \forall s \in \mathcal{S}_{\widetilde{\pi}}.
\end{equation}
The distillation of the states in $S_{\widetilde{\pi}}$ has similar effects for $\pi$. Encouraging $\pi$ to choose the actions from $\widetilde{\pi}$ for the states in $S_{\widetilde{\pi}}$ will push the values of $\pi$ towards optimal values because these actions lead to larger values based on $V^{\widetilde{\pi}}(\cdot)$. Thus, both Eq.~\ref{label:eq4} and \ref{label:eq5} will optimize the values of the two policies.

\section{Experiments}
\vspace{-0.05cm}
In this section, we empirically evaluate the proposed dual policy distillation~(DPD) framework. We develop two instances of DPD by using two DDPG learners and two PPO learners, and evaluate them on four continuous control tasks. Our experiments are designed to answer the following questions:
\begin{itemize}[wide=0pt, leftmargin=\dimexpr\labelwidth + 2\labelsep\relax]
\vspace{-3pt}
    \item \textbf{Q1:} Is DPD able to improve the performance in both on-policy and off-policy settings~(Sec.~\ref{sec:overall})?
    \item \textbf{Q2:} How will the values and actions outputted by the models evolve during training~(Sec.~\ref{sec:analysis})?
\end{itemize}

\begin{figure*}
  \centering
  \begin{subfigure}[b]{0.25\textwidth}
    \includegraphics[width=1.0\textwidth]{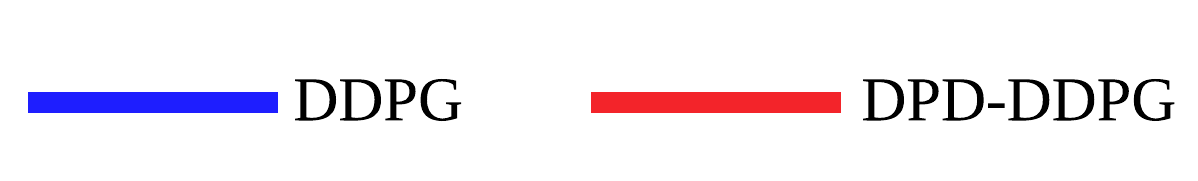}
  \end{subfigure}
  
  \begin{subfigure}[b]{0.25\textwidth}
    \includegraphics[width=0.95\textwidth]{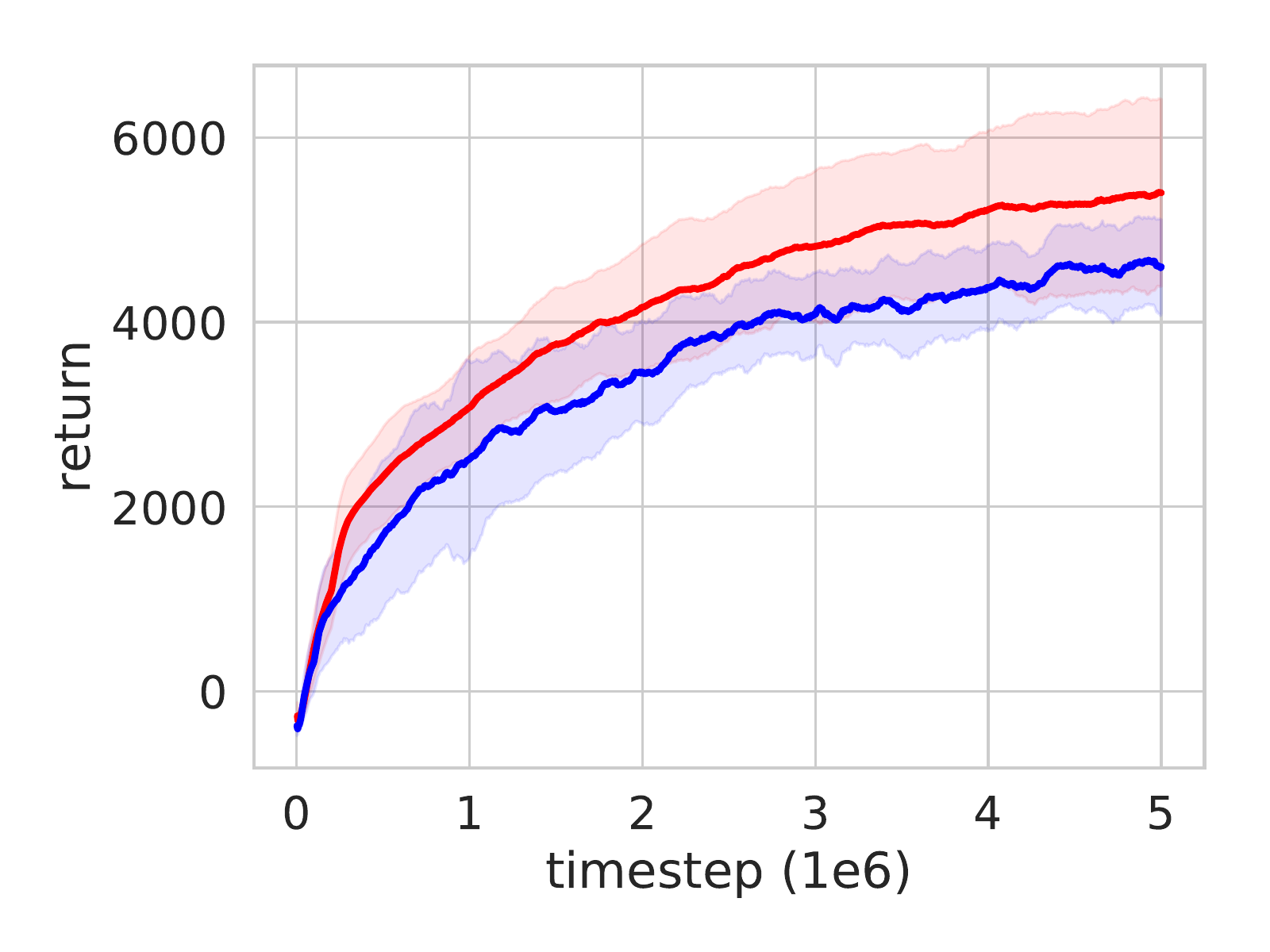}
    \subcaption{HalfCheetah}
  \end{subfigure}%
  \begin{subfigure}[b]{0.25\textwidth}
    \includegraphics[width=0.95\textwidth]{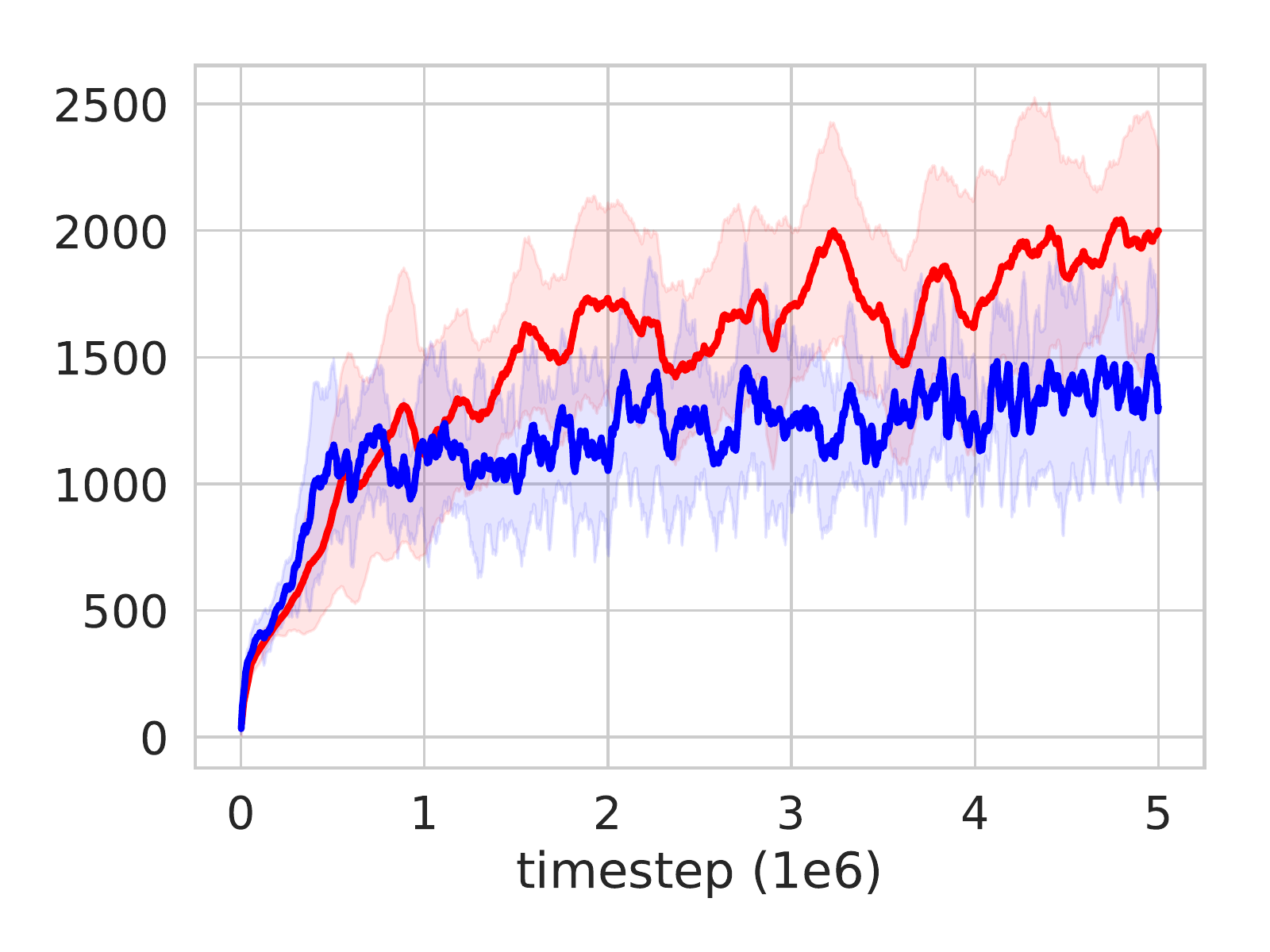}
    \subcaption{Walker2d}
  \end{subfigure}%
  \begin{subfigure}[b]{0.25\textwidth}
    \includegraphics[width=0.95\textwidth]{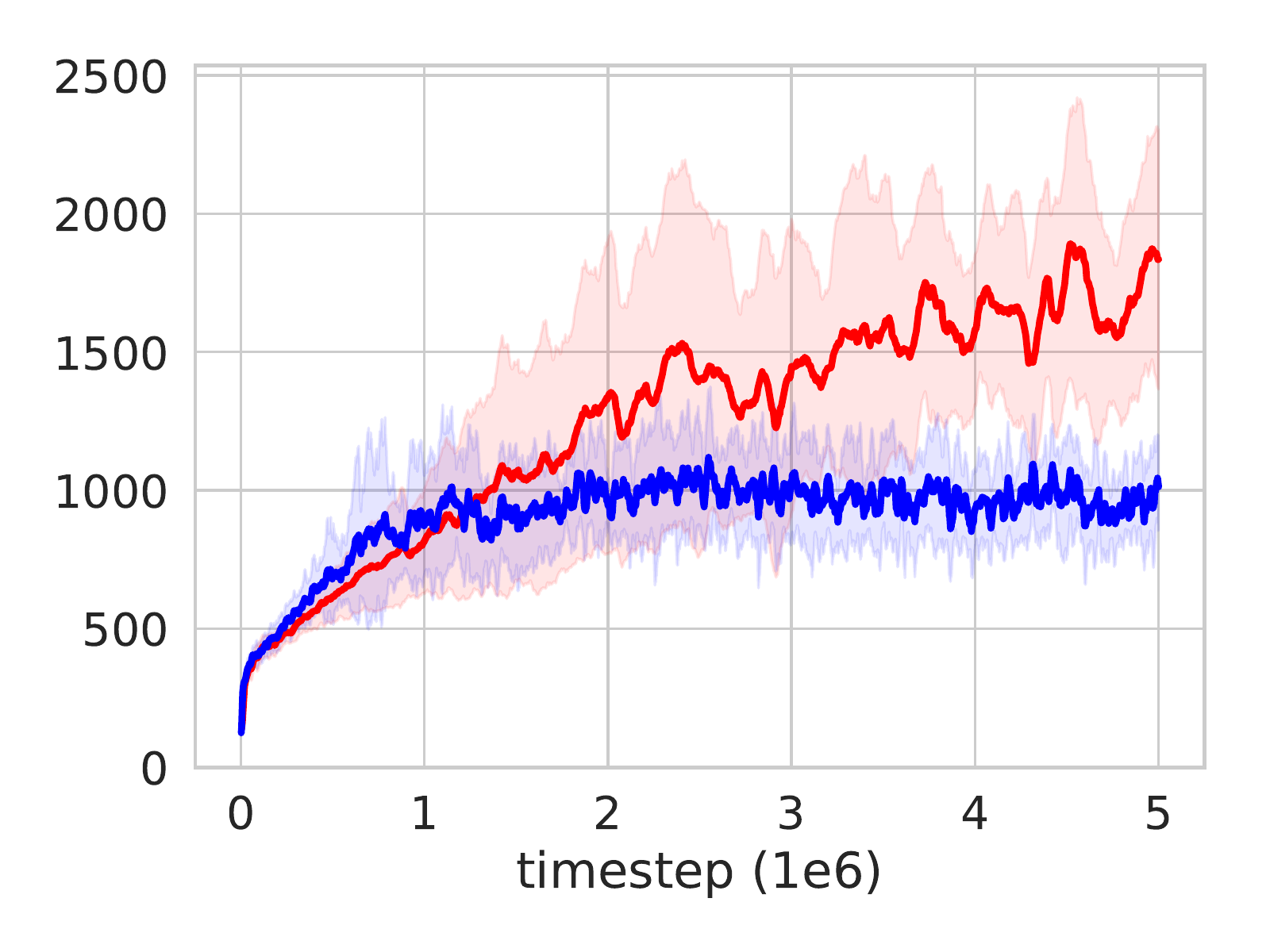}
    \subcaption{Humanoid}
  \end{subfigure}%
  \begin{subfigure}[b]{0.25\textwidth}
    \includegraphics[width=0.95\textwidth]{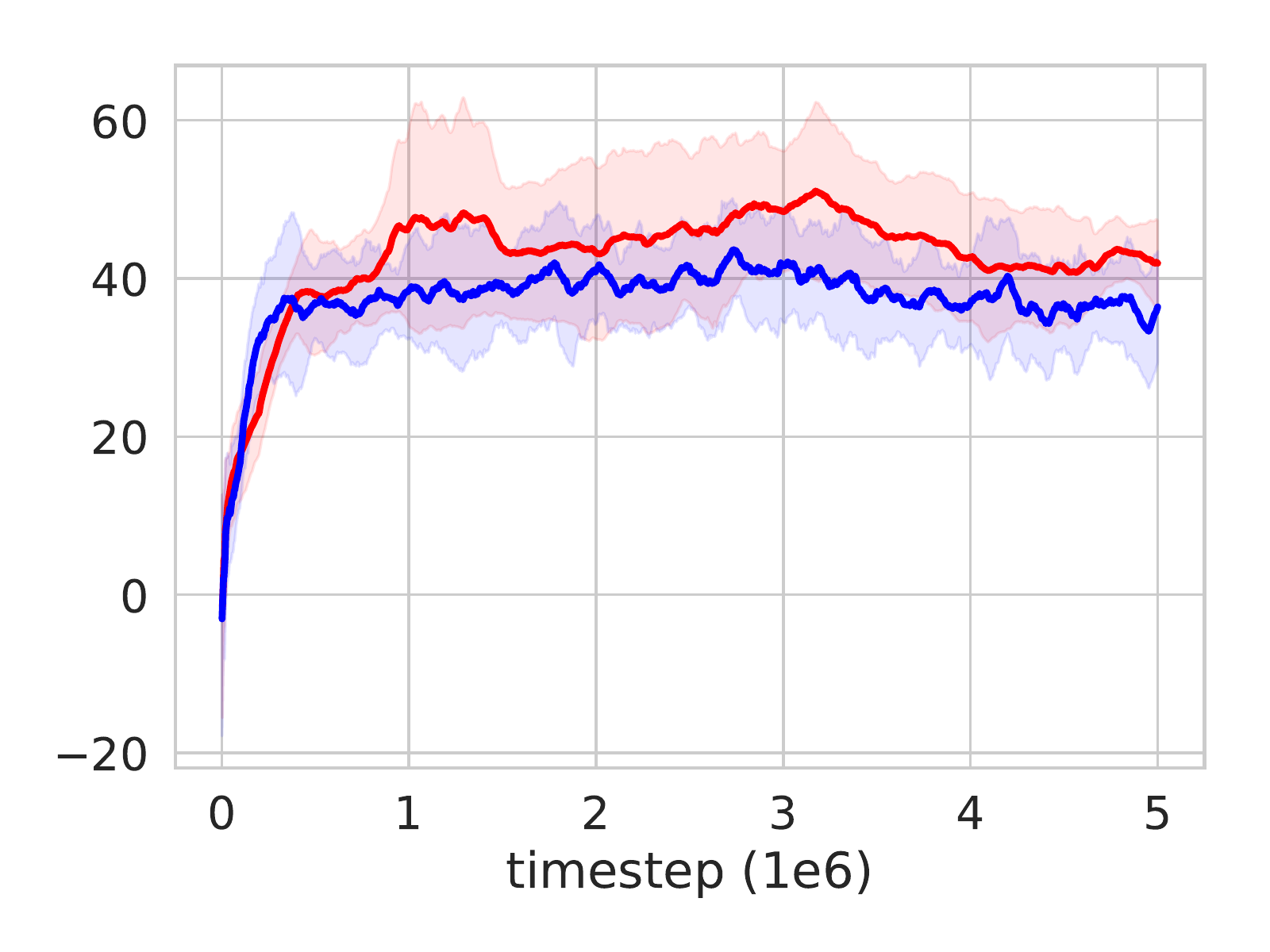}
    \subcaption{Swimmer}
  \end{subfigure}%
  \vspace{-8pt}
  \caption{Overall performance comparison on four continuous control tasks in off-policy setting. The shaded area represents mean $\pm$ standard deviation. For a fair comparison, each learner of DPD is run for $2.5 \times 10^6$ timesteps, and the X-axis (timestep) of DPD is stretched to $5.0 \times 10^6$ . Thus, all the algorithms are compared in the same condition with $5 \times 10^6$ timesteps in total. The learning curves are averaged over $10$ random seeds. The performance is measured by the average return over episodes.}
  \label{fig:main}
\end{figure*}
\vspace{-10pt}

\subsection{Experimental Setting}
\vspace{-0.05cm}
Our experiments are implemented upon PPO~\cite{schulman2017proximal} and DDPG~\cite{lillicrap2016continuous}, which are benchmark RL algorithms. We use the implementation from OpenAI baselines\footnote{\url{https://github.com/openai/baselines}} and follow all the hyper-parameters setting and network structures for our DPD implementation and all the baselines we considered.  Since the policy network in DDPG is deterministic, we directly use the outputs of the Q-network in DDPG to estimate the state values. We use mean square error to compute the distance between the output actions of the two policies for the distillation objective. We consider the following baselines:

\begin{itemize}[wide=0pt, leftmargin=\dimexpr\labelwidth + 2\labelsep\relax]
    \item \textbf{DDPG:} the vanilla DDPG without policy distillation.
    \item \textbf{PPO:} the vanilla PPO without policy distillation.
\vspace{-5pt}
\end{itemize}

The experiments are conducted on several continuous control tasks from OpenAI gym\footnote{\url{https://gym.openai.com/}}~\cite{brockman2016openai}: Swimmer-v2, HalfCheetah-v2, Walker2d-v2, Humanoid-v2.

\subsection{Implementation Detail}
\vspace{-0.05cm}
For the off-policy setting, we used an 64-64-64 MLP, a memory buffer with size $10^{6}$, \textit{RELU} as activation functions within layers and \textit{tangh} for the final layer. For the on-policy setting, we used 64-64 MLP for policy network, 64-64-64 MLP for value network, and \textit{tanh} as activation functions for both network. To facilitate the policy distillation, we use the transitions from the most recent 1000 transitions for policy distillation. The DPD algorithm has an additional parameter $\alpha$, which controls the confidence level of the estimated values. We empirically select its value for each setting from the range $[0.1, 10.0]$. The learning rate and the batch size for the distillation are set to $10^{-4}$ and $64$ for off-policy setting and $10^{-5}$ and $256$ for on-policy setting. 

\subsection{Overall Performance}
\vspace{-0.05cm}
\label{sec:overall}
We conduct experiments on both on- and off-policy settings to validate the proposed disadvantageous distillation strategy is beneficial in general.
We run as many timesteps and the number of trials as our computational resources allow. Since DPD requires updating two policies simultaneously, we run vanilla DDPG and PPO for twice the number of timesteps as that of DPD for a fair comparison, i.e., we run DPD-DDPG and DPD-PPO for $2.5 \times 10^6$ and $1 \times 10^7$ timesteps; and vanilla DDPG and PPO for $5 \times 10^6$ and $2 \times 10^7$ timesteps. Figure~\ref{fig:main} plot the learning curve of off-policy settings, and the result of on-policy settings is tabulated in Table~\ref{table:main}. 


Based on the result, we observed that, each of the two policies in DPD is significantly enhanced by the dual distillation, and DPD outperforms vanilla DDPG and PPO within the same running timesteps in the $4$ tasks. Specifically, when comparing DPD and DDPG at $2.5 \times 10^6$ timesteps, the maximum return of DPD has an improvement of more than $15$ percent in $3$ out of $4$ tasks.
By comparing the performance of DPD at $2.5 \times 10^6$ timesteps with that of both PPO and DDPG at $5 \times 10^6$ timesteps, the maximum return of DPD has an improvement of more than $10$ percent in the tasks. In our experiments, we only empirically explore $\alpha$ over a small set. It is possible to further improve DPD if exploring more $\alpha$ values. The results suggest that it is promising to exploit the knowledge from a peer policy and the proposed framework is generally applicable for both on- and off-policy algorithms.


\begin{figure*}
  \centering
  \begin{subfigure}[b]{0.40\textwidth}
    \includegraphics[width=\textwidth]{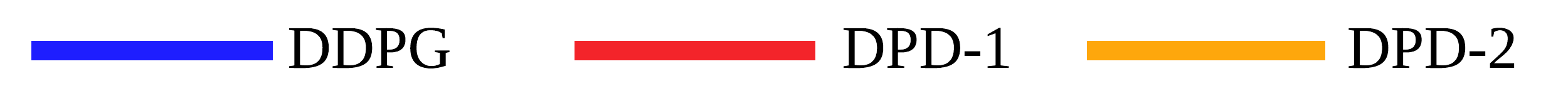}
  \end{subfigure}
  
  \begin{subfigure}[b]{0.25\textwidth}
    \includegraphics[width=0.75\textwidth]{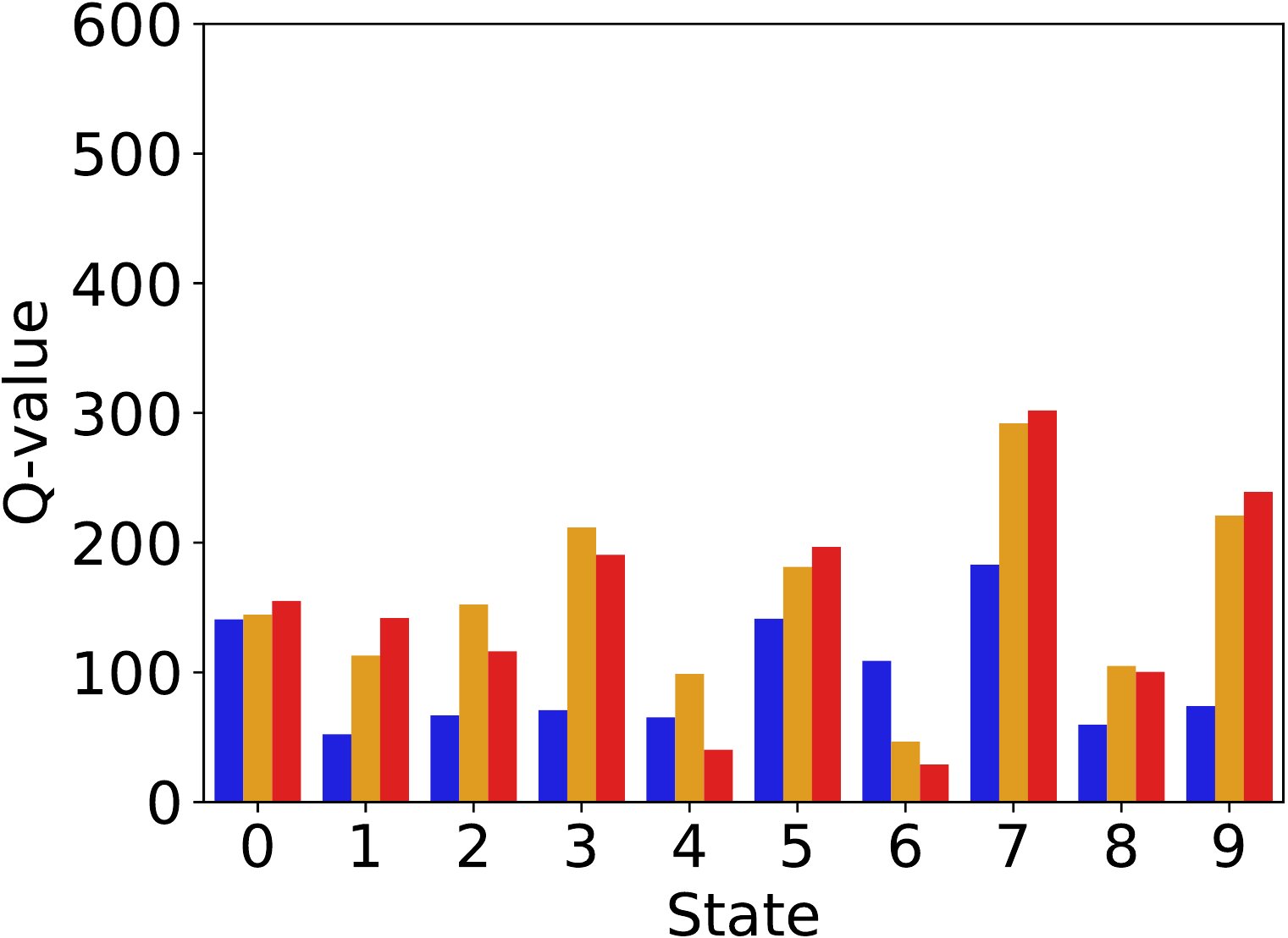}
    \subcaption{Early~(Q-value)}
  \end{subfigure}%
  \begin{subfigure}[b]{0.25\textwidth}
    \includegraphics[width=0.75\textwidth]{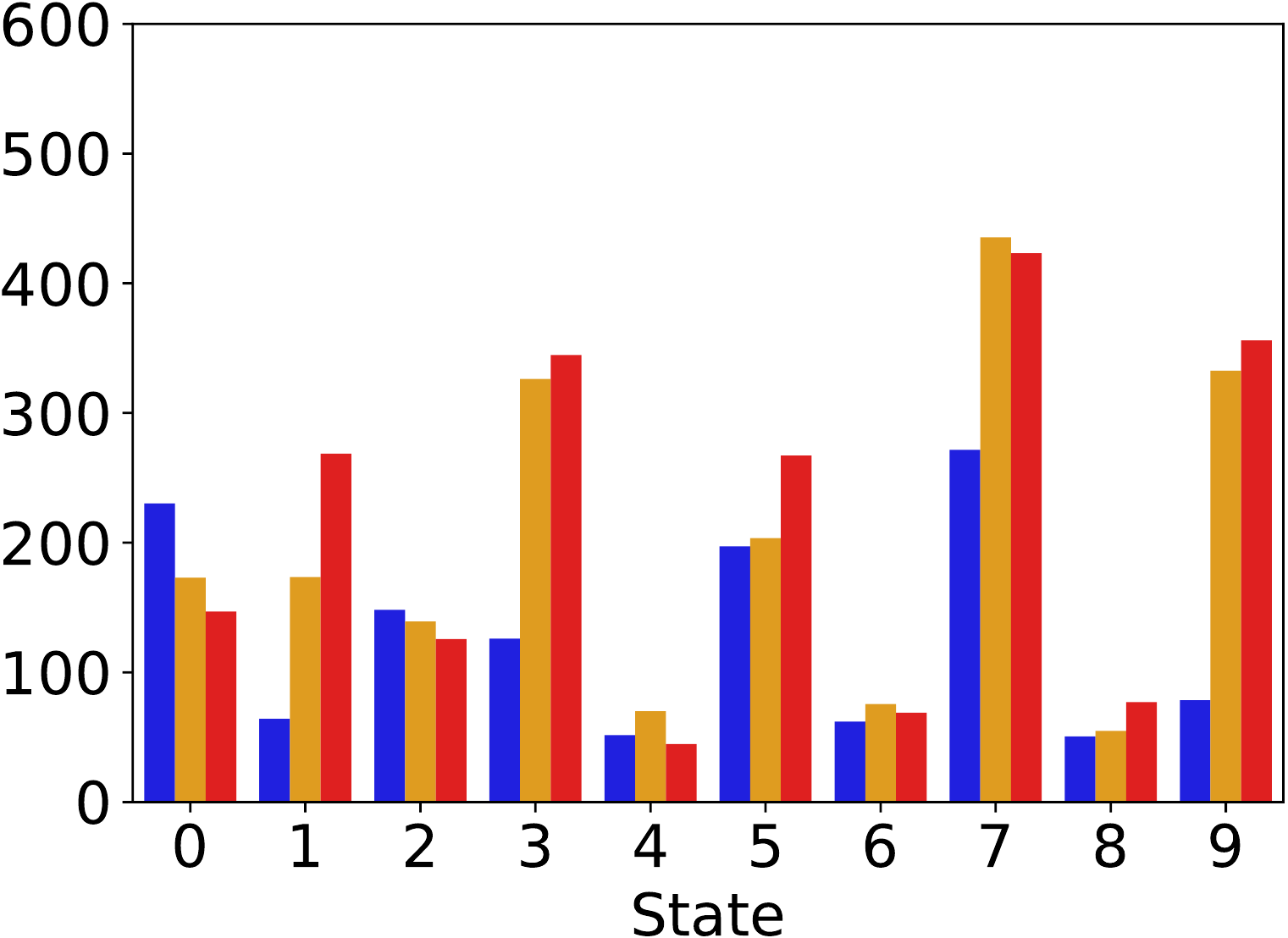}
     \subcaption{Middle~(Q-value)}
  \end{subfigure}%
  \begin{subfigure}[b]{0.25\textwidth}
    \includegraphics[width=0.75\textwidth]{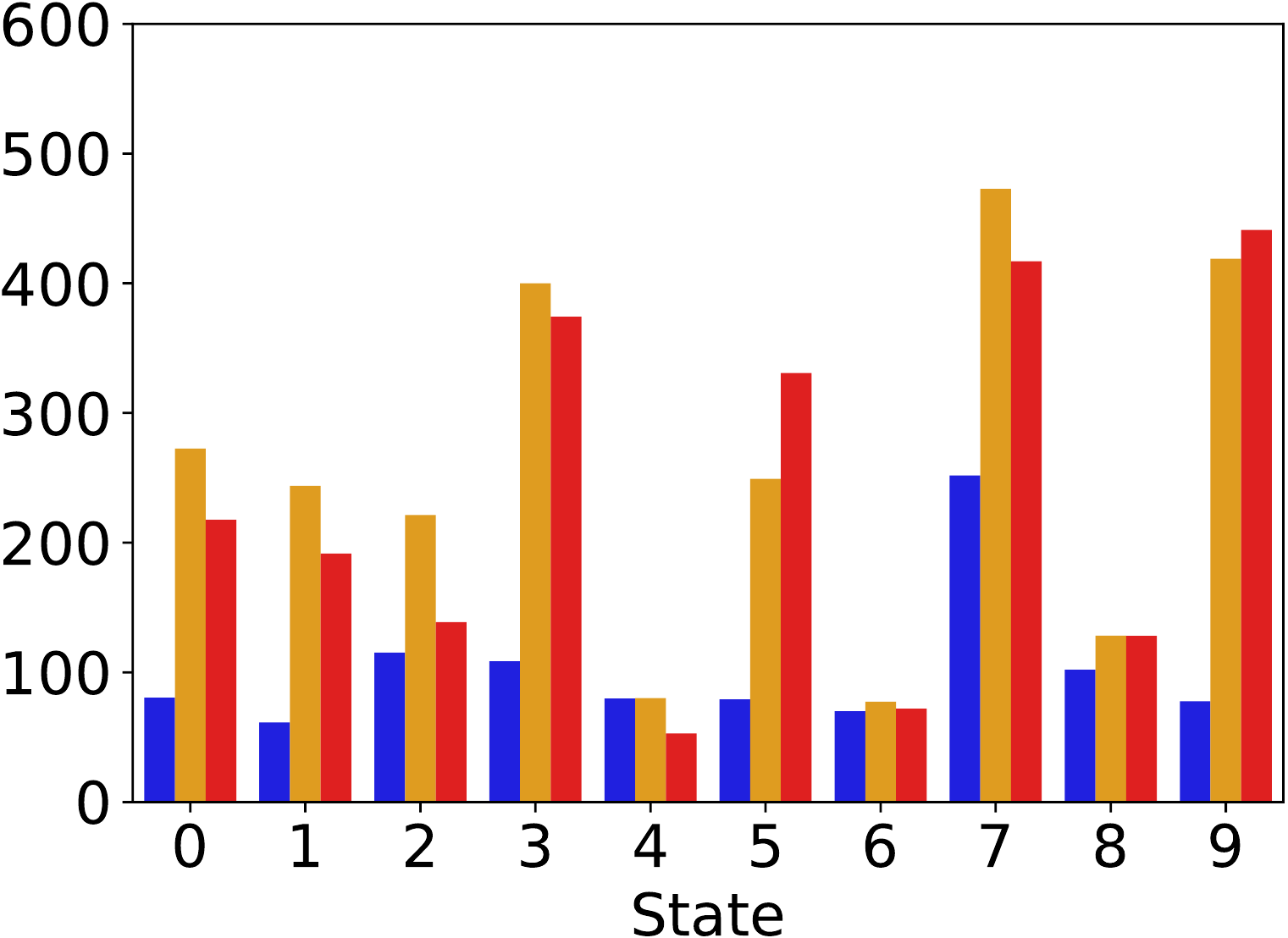}
     \subcaption{Late~(Q-value)}
  \end{subfigure}%
  
  \begin{subfigure}[b]{0.30\textwidth}
    \includegraphics[width=\textwidth]{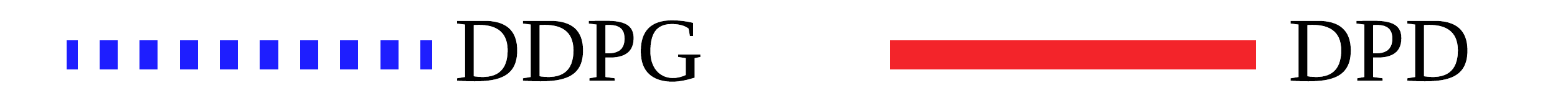}
  \end{subfigure}
  
  \begin{subfigure}[b]{0.25\textwidth}
    \includegraphics[width=0.75\textwidth]{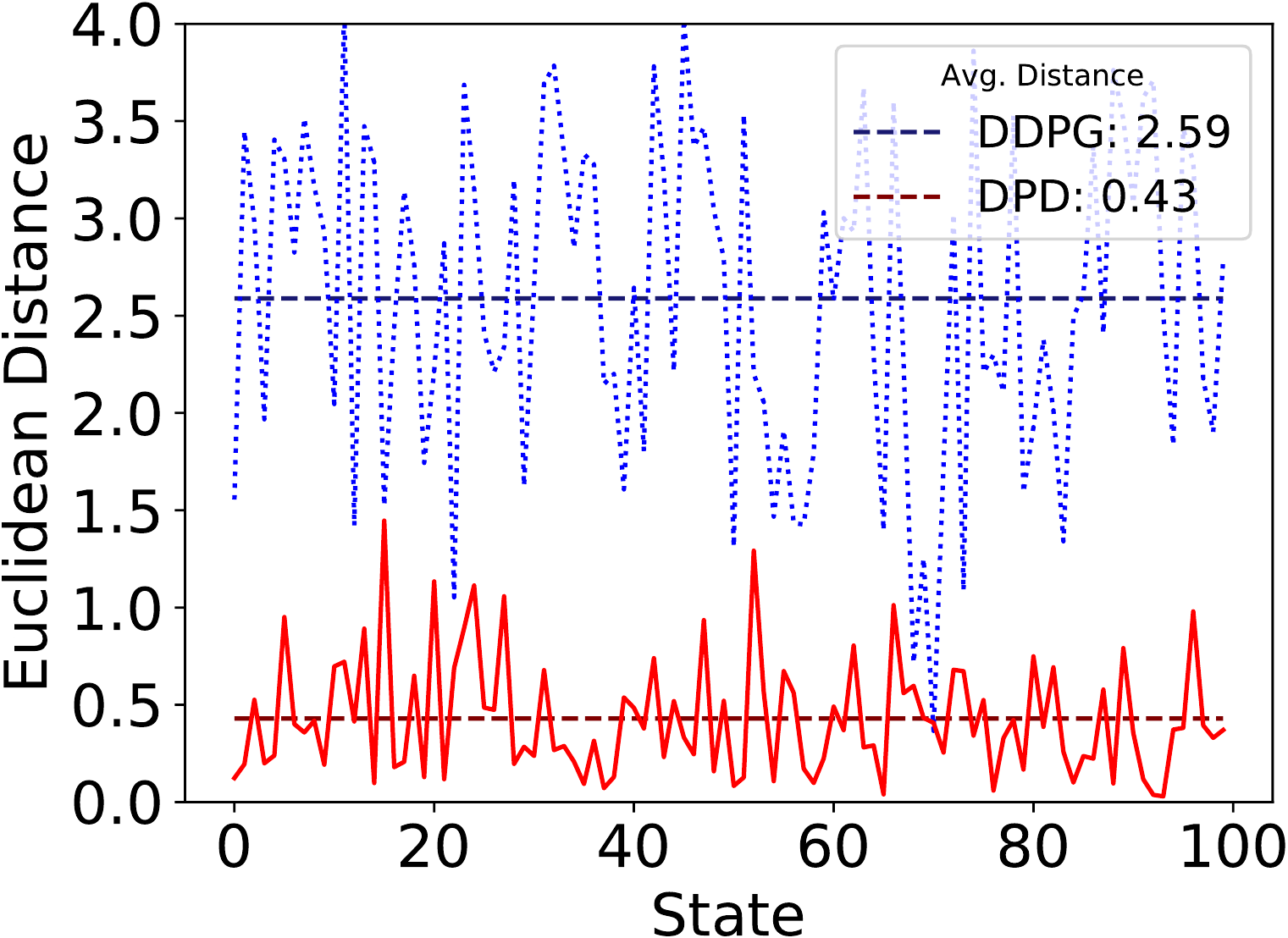}
    \subcaption{Early~(action)}
    \vspace{-1.5pt}
  \end{subfigure}%
  \begin{subfigure}[b]{0.25\textwidth}
    \includegraphics[width=0.75\textwidth]{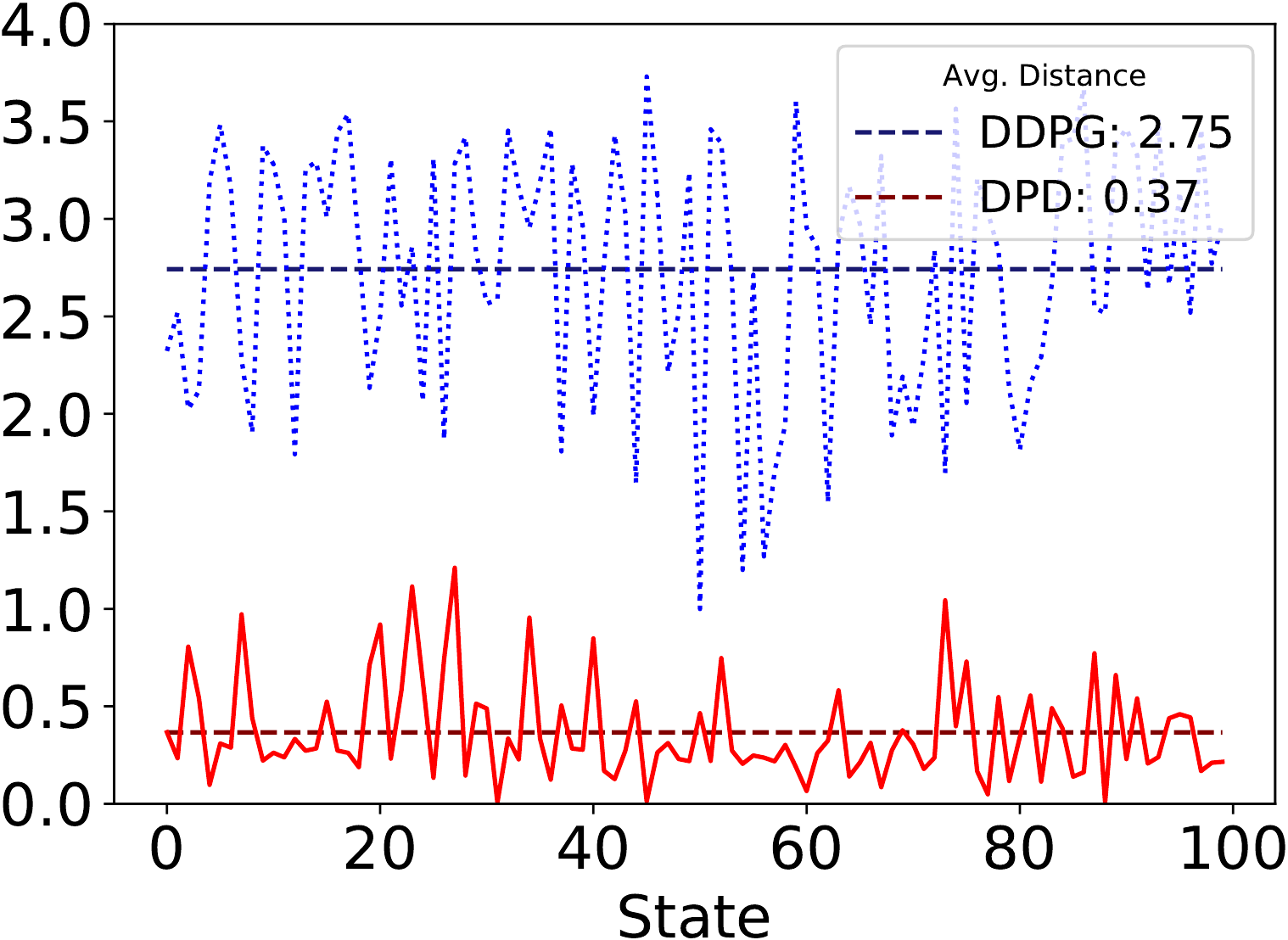}
     \subcaption{Middle~(action)}
     \vspace{-1.5pt}

  \end{subfigure}%
  \begin{subfigure}[b]{0.25\textwidth}
    \includegraphics[width=0.75\textwidth]{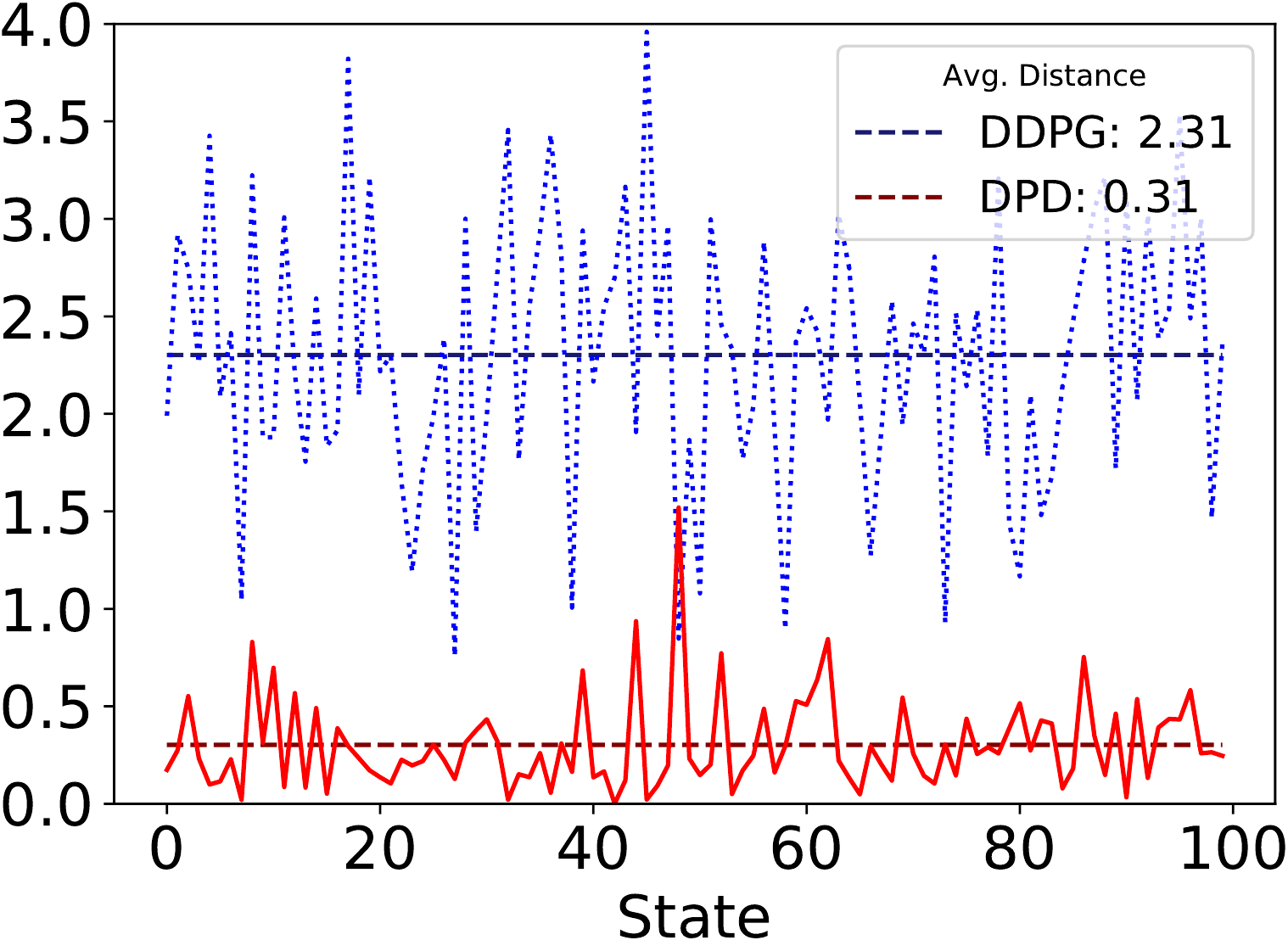}
     \subcaption{Late~(action)}
     \vspace{-1.5pt}

  \end{subfigure}%
  
  \caption{The evolution of Q-values~(top) and actions~(bottom) outputted by DPD and DDPG in Walker2d. Since DDPG uses a deterministic policy network, the Q-values can represent the state values. DPD-1 and DPD-2 denote the first learner and the second learner in DPD respectively. The dark lines are the average values of the two distances. The x-axis represents the state index. For Q-values evolution, we randomly sample $10$ states from the rollouts performed by a pre-trained model which is obtained by training DDPG for $5 \times 10^6$ timesteps. From left to right, we plot the Q-values of these $10$ states in early~($5 \times 10^5$ timesteps), middle~($1.5 \times 10^6$ timesteps) and later~($2.5 \times 10^6$ timesteps) training stages. For actions evolution, we run DDPG for two separate runs with different random seeds and compute the Euclidean distance between the actions outputted by the two DDPG models as well as the actions outputted by the two learners in DPD. We plot the results of $100$ randomly sampled states. Note that we sample $10$ or $100$ states for better visualization. We have sampled multiple times and observed similar results.}
  \label{fig:analysis}
  \vspace{-10pt}
\end{figure*}

\begin{table}[]
\centering
\begin{tabular}{@{}lll@{}}
\toprule
\textit{Mean} & PPO & DPD-PPO   \\ \midrule
HalfCheetah   & 2947.17 $\pm{201.11}$  &  \textbf{3051.28} $\boldsymbol{\pm{190.51}}$  \\
Walker2d      & 3694.79 $\pm{224.12}$ &  \textbf{3857.23} $\boldsymbol{\pm{143.58}}$ \\
Humanoid      & 2164.36 $\pm{127.40}$  & \textbf{2242.19} $\boldsymbol{\pm{230.72}}$ \\
Swimmer       & 97.36 $\pm{0.57}$   & \textbf{98.90} $\boldsymbol{\pm{2.50}}$  \\ \toprule
\textit{Max} & PPO & DPD-PPO   \\ \midrule
HalfCheetah   & 4031.77 $\pm{3048.85}$  &  \textbf{6373.40} $\boldsymbol{\pm{941.01}}$  \\
Walker2d      & 4738.16 $\pm{161.22}$  &  \textbf{5233.56} $\boldsymbol{\pm{286.84}}$  \\
Humanoid      & 3518.00 $\pm{278.31}$  & \textbf{3885.83} $\boldsymbol{\pm{261.89}}$ \\
Swimmer       & 110.82 $\pm{0.83}$  & \textbf{120.99} $\boldsymbol{\pm{8.01}}$ \\ \bottomrule
\end{tabular}

\caption{Overall performance comparison on four continuous control tasks in on-policy setting. For a fair comparison, each learner of DPD is run for $1 \times 10^7$ timesteps, and the PPO learner is run for $2 \times 10^7$ timesteps.
The result are averaged over $5$ random seeds. The performance is measured by the mean return over episodes and mean of maximum return of each episode.}
\label{table:main}
\vspace{-5pt}
\end{table}

\subsection{Analysis of the Dual Distillation}
\vspace{-0.05cm}
\label{sec:analysis}
To better understand the dual distillation mechanism, we study how the Q-values and the actions outputted by the two learners evolve under DPD framework. We randomly sample some states from the rollouts performed by a pre-trained model which is obtained by training DDPG for $5 \times 10^6$ timesteps. We then feed these states to the DPD and DDPG models in different training stages. The outputted Q-values and actions are illustrated in Figure~\ref{fig:analysis}~(we have sampled multiple times and observed similar results). We empirically find some insights.

For the Q-values, we make two observations. First, in all the training stages, the Q-values outputted by each of the learners tend to be larger at some states and smaller at some other states compared with the other learner. The result supports our hypothesis that the two learners are complimentary so that we can find a hypothetical hybrid policy that has guaranteed policy improvement, although it is possible that it is caused by inaccurate Q-values estimation. Second, we can see that the Q-values of all the states tend to increase throughout the training process, and they increase faster than those of DDPG. The result suggests that the proposed dual distillation indeed pushes the values of the two learners towards the larger optimal values, as indicated by our theoretical justification.

For the actions, we calculate the Euclidean distance between the outputted actions of the two learners in DPD. We also run two separate DDPG models and calculate the Euclidean distance between their outputted actions for comparison. As expected, we observe that in all training stages, the two learners in DPD tend to perform more similar actions than the two DDPG models. We can see that the average Euclidean distance for DPD decreases from the early stage~($0.43$) to the later stage~($0.31$), which suggests that the dual distillation may make the two learners slowly converge through the training process. Since the two learners have similar policies, the result further supports our assumption that the two learners in DPD have similar state visiting frequencies.

\vspace{-5.0pt}
\section{Related Work}
\vspace{-7.0pt}
\paratitle{Teacher-student framework}
A typical teacher-student framework is imitation learning.~\cite{abbeel2004apprenticeship,finn2016guided}. One representative method is behavior cloning~\cite{bain1999framework,ho2016generative,rusu2015policy}, which directly trains a student policy based on demonstration data of experts. 
Previous work shows that it is promising to learn from imperfect demonstrations~\cite{hester2018deep} and exploiting own past good experiences will help exploration~\cite{oh2018self}. Our framework also learns from imperfect demonstrations, but treats actions performed by a peer policy as demonstrations. Collaborative learning is also studied in~\cite{lin2017collaborative}, however, with expensive pre-trained teachers. Meta-learning methods also make use of a teacher model to improve the sample efficiency~\cite{xu2018meta,xu2018learning,zha2019experience}. Our work extends the traditional teacher-student setting and studies a student-student framework with two student models distilling knowledge from each other. Each model servers as both student and teacher and work together with its peer model to find the solution.
\vspace{-5pt}

\paratitle{Multi-agent reinforcement learning} Multi-agent reinforcement learning~\cite{littman1994markov,tan1993multi,shoham2003multi} studies how a group of agents sharing the same environment learns to collaborate and coordinate with each other to achieve a goal. There are several studies on knowledge diffusion in multi-agent setting~\cite{hadfield2016cooperative,da2017simultaneously,omidshafiei2018learning}. Although our work also introduces two learners, our setting is significantly different from the multi-agent setting in that the two learners in the framework are independently deployed to two instances of the same single-agent environment. The two learners capture various aspects of the same environment and share beneficial knowledge during training.

\section{Conclusion and Future Work}
\vspace{-10pt}
In this work, we introduce dual policy distillation, a student-student framework which enables two policies to explore different aspects of the environment and exploit the knowledge from each other. Specifically, we propose a distillation strategy which prioritizes the distillation at disadvantage states from the peer policy. We theoretically and empirically show that the proposed framework can enhance the policies significantly. There are two interesting future directions. First, we will investigate whether our framework can be extended to enable knowledge transfer among multiple tasks. Second, we would like to explore the possibility of using our framework to combine the benefits of different RL algorithms. 
\section*{Acknowledgements}
This work is, in part, supported by NSF (\#IIS-1750074, \#IIS-1939716 and \#IIS-1900990). The views and conclusions contained in this paper are those of the authors and should not be interpreted as representing any funding agencies.
\bibliographystyle{named}
\bibliography{ijcai20}

\end{document}